\documentclass[preprint,12pt,number]{elsarticle}

\usepackage{amsmath,amssymb}
\usepackage{graphicx}
\usepackage{booktabs}
\usepackage{array}
\usepackage{adjustbox}
\usepackage{multirow}
\usepackage[section]{placeins}
\usepackage{xcolor}
\usepackage{xurl}
\usepackage{hyperref}
\hypersetup{colorlinks=true,linkcolor=black,citecolor=black,urlcolor=blue}
\biboptions{sort&compress}

\graphicspath{{figures/}}

\journal{Engineering Applications of Artificial Intelligence}

\begin{document}

\begin{frontmatter}

\title{Unsupervised detection of underground tunnels in ground-\hspace{0pt}penetrating radar using depth-restricted reconstruction scoring\tnoteref{abbr}}
\tnotetext[abbr]{Abbreviations: AP, average precision; AUC, area under the receiver operating characteristic curve; CI, confidence interval; CPU, central processing unit; FP, false positive; FPR, false-positive rate; GPR, ground-penetrating radar; GPU, graphics processing unit; NARC, National Agricultural Research Centre; OCSVM, one-class support vector machine; PaDiM, patch distribution modeling; PCA, principal component analysis; ReLU, rectified linear unit; ROC, receiver operating characteristic; SD, standard deviation; SEC, spreading and exponential compensation.}

\author[case]{Muhammad Junaid\corref{cor1}}
\ead{junaid6330@yahoo.com}
\author[case]{Shoab A. Khan}
\ead{kshoab@yahoo.com}
\author[strath]{Nisar Ahmed}
\ead{ch.nisar89@icloud.com}

\cortext[cor1]{Corresponding author.}

\affiliation[case]{organization={Department of Electrical Engineering \& Computer Science, Sir Syed CASE Institute of Technology},
            city={Islamabad},
            country={Pakistan}}
\affiliation[strath]{organization={Department of Computer \& Information Sciences, University of Strathclyde},
            city={Glasgow},
            country={United Kingdom}}

\begin{abstract}
Hand-dug tunnels that reach buried pipelines from below are hard to find before they are used, and too rare to train a supervised detector. The engineering application of this work is detecting such tunnels in ground-penetrating radar surveys of a monitored site. The contribution in artificial intelligence is a detection scheme that needs no tunnel examples: a convolutional autoencoder learns the site's normal ground from surveys recorded before any tunnel exists, later survey windows are scored by their largest reconstruction errors within the travel-time band where a tunnel can lie, and evaluation holds out whole survey lines. On field data from a tunnel hand-dug for this study (0.6\,m wide, 1\,m high, roof 0.9--1.4\,m deep) and surveyed repeatedly with a 250-megahertz radar in dry and wet conditions, the detector reaches an area under the receiver operating characteristic curve of 0.994 and an F1 score of 0.972 on 1,600 test windows and flags all 29 tunnel crossings. Restricting the score to the depth band improves all six detectors tested and, with a quarter of the training data, outperforms whole-window scoring trained on all of it; the best rule for the same band without learning misses 9 of the 29 crossings. On two unseen plots, false alarms stay at 1.6\% on comparable ground but reach 9.3\% on irrigated soil; recalibrating the threshold on a fifth of that plot's lines lowers them to 4.2\%, retraining brings no significant further reduction, and models trained there alone still detect the original tunnel.
\end{abstract}

\begin{highlights}
\item Tunnels found in radar surveys by a model trained only on pre-tunnel ground
\item Scoring only the depth band where tunnels lie improves all six detectors
\item All 29 crossings of a hand-dug tunnel detected; misses only at crossing edges
\item Three tunnels hand-dug and surveyed repeatedly to build a public field dataset
\item Trained at one site, it found the tunnel at another after resetting its alarm level
\end{highlights}


\end{frontmatter}

\section{Introduction}
\label{sec:intro}

Buried pipelines carry oil and gas across long, sparsely monitored corridors, and their value makes them a recurring target of unauthorised underground access. In January 2024, for example, customs officers in Karachi, Pakistan, uncovered a 174-foot tunnel connected to a fuel pipeline and used to divert diesel \cite{profit2024}. Conventional monitoring---pressure sensing, flow balancing and surface patrols---generally reacts to a breach that has already occurred and gives little warning of excavation approaching a pipeline from below \cite{oktenli2026}.

Ground-penetrating radar (GPR) can image this threat directly. An air-filled void presents a strong dielectric contrast to the surrounding soil and produces the hyperbolic reflections that analysts use to identify voids and tunnels \cite{daniels2004}. The obstacle is interpretation: surveys produce thousands of radargrams affected by soil heterogeneity, moisture and clutter, and manual review is slow and subjective \cite{oguntoye2023,liu2025review}.

Deep learning has improved automated GPR interpretation, but nearly all reported systems are supervised and need labelled examples of the target, typically pipes and other utilities \cite{liu2023,jafuno2025}. For tunnels this requirement is hard to meet: intrusion tunnels are rare, their geometry varies, and no public collection of tunnel radargrams exists at a useful scale. A practical detector should instead learn what the ground of a monitored site normally looks like and flag departures from it.

We study this setting with field data collected for the purpose. Because no radar data of a known tunnel were available, we built the test case ourselves: after trials of the radar at six locations across Islamabad, three tunnels were hand-dug on agricultural land, and one site was surveyed repeatedly with GPR before and after its tunnel was dug, under changing moisture conditions. A convolutional autoencoder was trained only on the pre-construction surveys and used to score windows from the later ones. The paper makes four contributions:
\begin{enumerate}
    \item \textbf{Depth-restricted reconstruction scoring.} Pooling the largest reconstruction errors only within the travel-time band in which a tunnel can occur improves every unsupervised detector we tested, compensates for a fourfold reduction in training data, and is robust to where the band starts. Scores of the same band that involve no learning detect the tunnel far less reliably, so the learned model of normal ground remains essential.
    \item \textbf{A line-held-out evaluation protocol for overlapping GPR windows}: thresholds set on other survey lines, confidence intervals from resampling whole lines, detection counted per tunnel crossing, and false alarms measured on ground never used for training.
    \item \textbf{Evidence for a train-once, calibrate-per-site deployment.} On an unseen, irrigated plot, recalibrating the threshold from a short baseline survey more than halves the false alarms, while retraining the model on that plot's ground gives no significant further reduction across all 136 of its survey lines.
    \item \textbf{A public field dataset} of 250-MHz GPR windows from a site surveyed before and after the construction of a hand-dug tunnel, together with normal ground from two further plots, released with all code. To our knowledge, no public GPR dataset of a purpose-built tunnel surveyed before and after its construction has been available.
\end{enumerate}
No tunnel example is used for training. Labels enter only the evaluation and the choice of normal windows from which thresholds are calibrated, which corresponds to calibrating on survey stretches known to be clean (Section~\ref{sec:protocol}).

\section{Related work}
\label{sec:related}

\subsection{Automated GPR interpretation}
Most automated GPR analysis targets buried utilities. Classical methods relied on hand-crafted descriptors of hyperbolic reflections followed by shallow classifiers \cite{paul2025}, an approach that needs signal-processing expertise and transfers poorly between sites. Convolutional networks removed the feature-engineering step, for example to recognise and localise pipelines in B-scans \cite{liu2023} and to classify buried objects \cite{jafuno2025}. Recent reviews catalogue this progress and its dependence on annotated data \cite{zou2025,liu2025review}. Other work improves the data themselves, for example by hybrid filtering of radargrams \cite{afrasiabi2025}. These systems are supervised and concentrate on pipes and cables rather than clandestine tunnels, whose scale, depth and rarity pose a different problem \cite{oktenli2026}. Supervised detectors can also be trained on simulated radargrams, for example from finite-difference time-domain modelling with gprMax \cite{warren2016}, but transferring them to field data requires the simulation to reproduce the heterogeneity and clutter of real soil; the approach studied here needs neither simulation nor tunnel examples.

\subsection{Unsupervised anomaly detection}
Reconstruction-based anomaly detection trains a model to reproduce normal data and uses reconstruction error as the anomaly score \cite{chandola2009,ruff2021}; denoising training \cite{vincent2008} discourages an identity mapping. In GPR, reconstruction-loss detection has been applied to radargram images \cite{hoang2024} and to pipe inspection \cite{angelis2024}. In a related field setting, an autoencoder trained only on background radiation spectra flagged radioactive sources and needed recalibration when the background changed \cite{sayre2026}. Classical alternatives model the normal data distribution directly, for example with principal component analysis (PCA) \cite{jolliffe2002}, one-class support vector machines (OCSVM) \cite{scholkopf2001} or isolation forests \cite{liu2008}. In industrial inspection, methods built on ImageNet-pretrained features---PaDiM \cite{defard2021} and PatchCore \cite{roth2022}---are current reference points. What this literature leaves open, and what we address, is how an anomaly score should be read from an error map when the target can only appear in a known, physically constrained part of the image, how such detectors should be evaluated when neighbouring windows overlap almost completely, and what is needed when a trained detector is moved to new ground.

\section{Field site and data}
\label{sec:data}

\subsection{Site and tunnel construction}

\begin{figure}[!htbp]
  \centering
  \includegraphics[width=\textwidth]{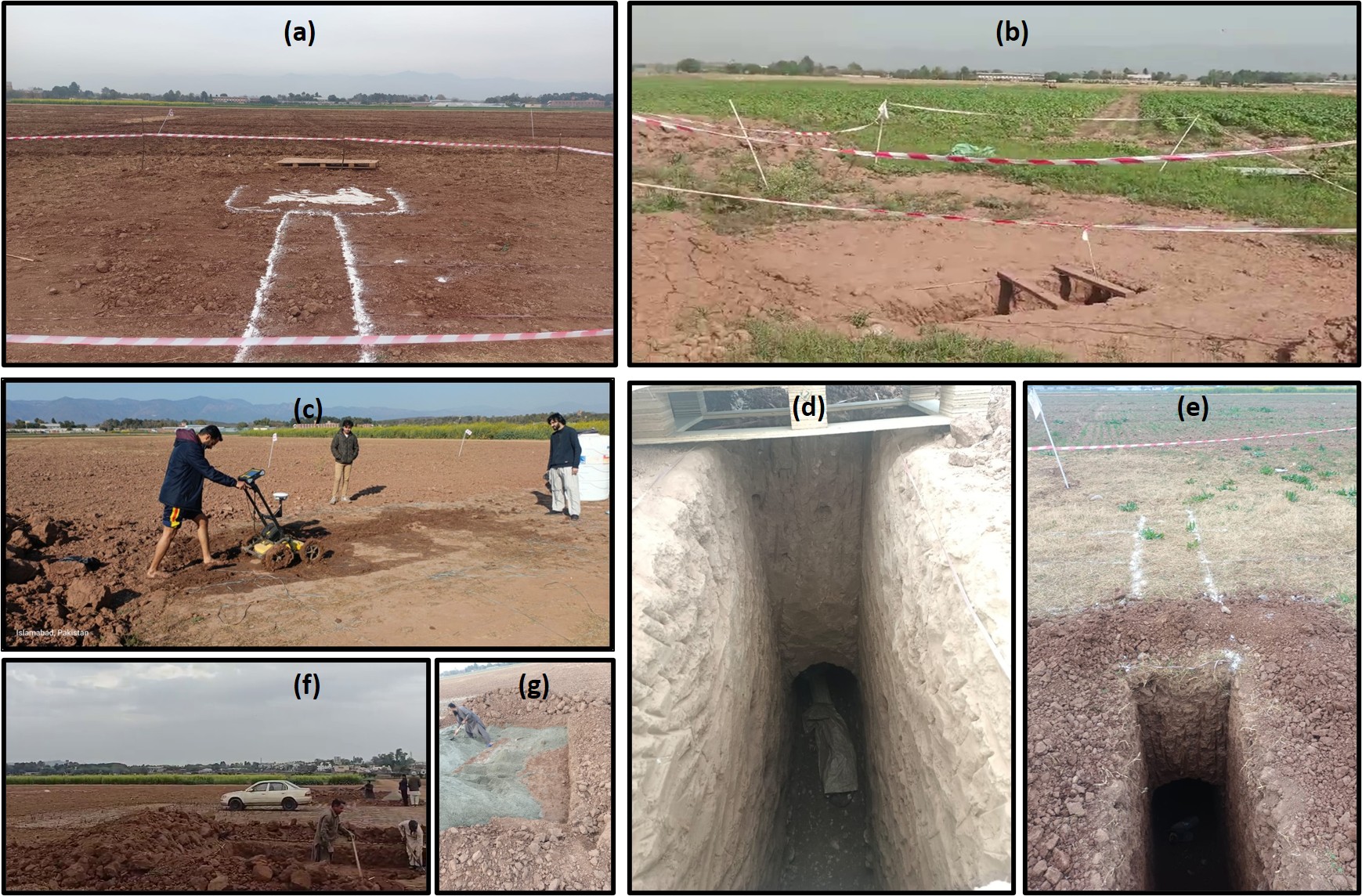}
  \caption{Field site on the land of the National Agricultural Research Centre, Islamabad, Pakistan. (a)~Route of the tunnel marked on the surface before tunnelling began. (b)~Entry shaft at the edge of the marked survey area. (c)~Data acquisition with the 250-MHz GPR along a marked survey line. (d)~View down the entry shaft into the tunnel. (e)~Entry shaft from the surface. (f), (g)~Excavation.}
  \label{fig:site}
\end{figure}

The study was carried out on agricultural land of the National Agricultural Research Centre (NARC), Islamabad, Pakistan (Fig.~\ref{fig:site}), where a published field study describes the soil as a loam of the Nabipur series \cite{akhtar2010}. At the main site (Site~A) the same strips of ground were surveyed repeatedly between January and March 2025, first before and then after a tunnel was dug beneath them, in dry conditions and after rain, by day and by night. The tunnel was hand-dug by trench-warfare-style tunnelling from a vertical entry shaft, leaving the ground surface above undisturbed. It has a roughly arched cross-section, about 0.6\,m wide and about 1\,m high (just large enough for one person to sit and dig), is more than 4.6\,m long, and is air-filled and unlined. Radar travel times place its roof about 0.9--1.4\,m below the surface, so its floor lies roughly 2\,m deep. Survey lines crossed the tunnel at right angles. Normal ground was also surveyed at two further plots on the same land (Sites~B and C, within about 300\,m of Site~A). Tunnels of the same type, but deeper (roofs at about 1.5 and 2\,m) and slightly narrower, were dug at these plots as well; their post-construction scans showed no clearly interpretable tunnel hyperbolas, so all tunnel data in this paper come from Site~A and Sites~B and C serve as normal ground only. The soil at Site~B was wet: the first author had previously irrigated it from two nearby water tanks for a separate soil-moisture study.

\subsection{Field campaign}
\label{sec:campaign}
No radar data of a known tunnel were available, so the test case was built for this study (Table~\ref{tab:campaign}). Between January and April 2025, the radar was first tried at six locations across Islamabad with different soils, over buried pipes, sewers and tunnels being dug for underground pipes and cables at construction sites, to judge how deep it could see and which ground suited it. Trial excavations on the NARC land were refilled with a different soil, such as sand, to observe the radar's response (Fig.~\ref{fig:site}f, g). After these trials, the NARC loam was chosen as the final site, and three tunnels were dug by hand, one per plot, each by three people in about a week. Over the campaign the radar was used on at least 13 field days and recorded more than 9\,km of GPS-tracked survey lines.

\begin{table}[!ht]
\caption{Field campaign at a glance.}
\label{tab:campaign}
\centering
\small
\begin{tabular}{>{\raggedright\arraybackslash}p{0.17\textwidth}>{\raggedright\arraybackslash}p{0.47\textwidth}>{\raggedright\arraybackslash}p{0.26\textwidth}}
\toprule
Stage & What was done & Extent \\
\midrule
Radar trials & Surveys over buried pipes, sewers and utility tunnels under construction, on different soils & Six locations across Islamabad \\
Trial excavations & Excavations on the NARC land refilled with a different soil, such as sand & Fig.~\ref{fig:site}f, g \\
Tunnel construction & Three tunnels hand-dug from vertical entry shafts, one per plot & About one week each, by three people \\
Surveys & Repeated passes in dry conditions and after rain, by day and by night, each with its GPS track & At least 13 field days; more than 9\,km of survey lines \\
Labelling & Labels assigned by the first author, checked by a second author and against the GPS tracks & 634 tunnel windows in 29 crossings \\
Dataset & Released publicly with all code & 20,954 windows from three plots \\
\bottomrule
\end{tabular}
\end{table}

\subsection{Acquisition}
Data were acquired with a Sensors \& Software LMX200 system with a 250-MHz antenna (0.25\,m antenna separation). Traces were triggered by an odometer every 0.05\,m and recorded with 500 samples per trace over a 200-ns time window, with 13--16 stacks per trace. The radar's built-in GPS receiver logged the track of every survey. Hyperbola fitting on tunnel reflections in EKKO\_Project gave a propagation velocity of about 0.085\,m/ns (relative permittivity $\approx 12.4$), which is used for all depth conversions in this paper.

\subsection{Processing and window extraction}
Radargrams were processed in EKKO\_Project with dewow filtering, background subtraction (filter width 5.05\,m) and SEC2 gain (spreading and exponential compensation; maximum gain 1000). Two gain settings were used, chosen per survey line for the clearest display: attenuation 32 and start gain 10 with the DynaT filter, or attenuation 15 and start gain 7 without it. Every window of a given line shares the same setting.

Windows were captured from the EKKO\_Project display. For each screen, the axes were removed and the leftmost square of the radargram was resized to $128{\times}128$ pixels. Each window therefore covers about 2.45\,m of survey line and the first 40\,ns of travel time (about 1.7\,m depth), so one pixel corresponds to about 1.9\,cm by 0.31\,ns. The display was advanced by a fixed scroll step between captures. Measured from the images, consecutive windows are offset by about 8 pixels (0.15\,m) on 28 survey lines and about 4 pixels (0.075\,m) on 27 lines, so neighbouring windows overlap by 94--97\%. A window was labelled \emph{tunnel} if the tunnel hyperbola was visible in it. Labels were assigned by the first author and checked independently by a second author (N.A.); disagreements were resolved by discussion. In addition, the location of each tunnel reflection was checked against the known position of the tunnel by viewing the radar's GPS tracks, exported from EKKO\_Project, in Google Earth.

\subsection{Dataset composition and integrity checks}
Table~\ref{tab:dataset} summarises the data. The 8,496 pre-construction windows from Site~A were split at random into training (80\%) and validation (20\%) sets. The test set contains 1,600 windows from 55 post-construction survey lines at Site~A (966 normal, 634 tunnel), kept in their spatial order; the tunnel windows form 29 crossings of the tunnel. Lines~1 and 2 are two segments of one scan and are treated as a single line wherever lines are grouped. Sites~B and C contribute 5,313 and 5,545 normal windows that were never used to train the published model.

No test window and no Site~B or C window is an exact copy of a training or validation window. To check how close the test windows are to the training data, each window was compared with its most similar training window (correlation at $32{\times}32$ resolution). The median correlation is 0.52 for normal and 0.46 for tunnel test windows, the maximum is 0.88, and no window exceeds 0.95: although the same strips were surveyed, changing conditions make every survey look different.

\begin{table}[!htbp]
\caption{Dataset composition.}
\label{tab:dataset}
\centering
\begin{adjustbox}{max width=\textwidth}
\begin{tabular}{llr}
\toprule
Use & Data & Windows \\
\midrule
Training & Site A, before tunnel construction (80\%) & 6,797 \\
Validation & Site A, before tunnel construction (20\%) & 1,699 \\
Test & Site A, after construction --- normal & 966 \\
Test & Site A, after construction --- tunnel (29 crossings) & 634 \\
Unseen ground & Site B, normal (irrigated soil; 136 survey lines) & 5,313 \\
Unseen ground & Site C, normal & 5,545 \\
\bottomrule
\end{tabular}
\end{adjustbox}
\end{table}

\section{Method}
\label{sec:method}

\subsection{Convolutional autoencoder}
\begin{figure}[!htbp]
\centering
\includegraphics[width=0.9\textwidth]{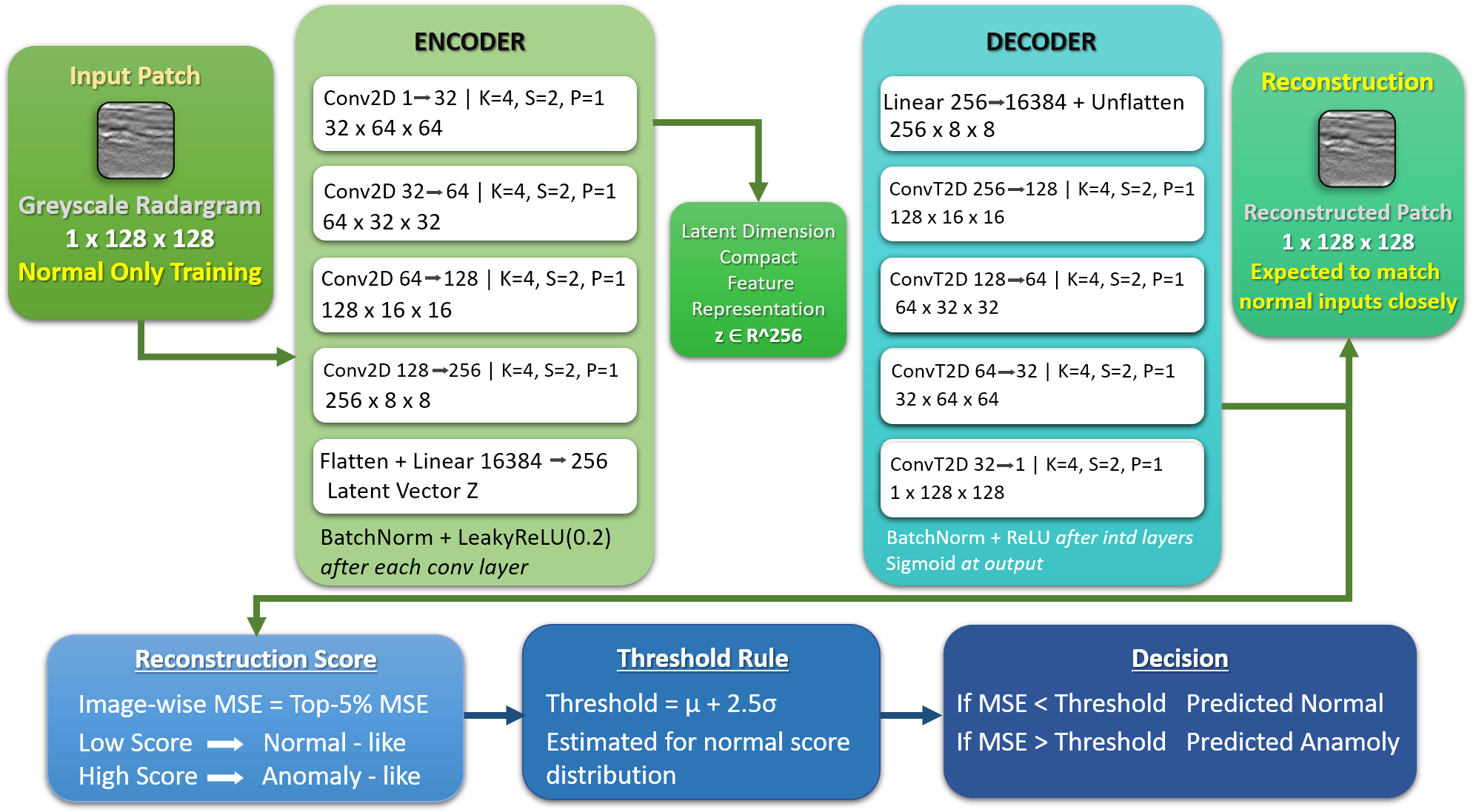}
\caption{Detection pipeline. The encoder compresses each $128{\times}128$ window to a 256-dimensional code and the decoder reconstructs it. The anomaly score averages the largest 5\% of squared reconstruction errors; in the proposed variant only rows 64--127 (the depth band, Section~\ref{sec:depth}) are eligible. A window is flagged when its score exceeds $\mu + 2.5\sigma$ of normal-window scores.}
\label{fig:arch}
\end{figure}

The detector is a convolutional bottleneck autoencoder (Fig.~\ref{fig:arch}, Table~\ref{tab:arch}). The encoder compresses the $1{\times}128{\times}128$ input through four convolutional blocks (kernel $4{\times}4$, stride 2, padding 1; 32, 64, 128 and 256 channels), each followed by batch normalisation and LeakyReLU (slope 0.2). The resulting $256{\times}8{\times}8$ map is flattened and projected to a 256-dimensional code. The decoder mirrors the encoder with transposed convolutions, batch normalisation and ReLU, and ends with a sigmoid. The model has 9.78 million parameters.

\begin{table}[!htbp]
\caption{Autoencoder architecture. Conv2D: two-dimensional convolution (kernel $4{\times}4$, stride 2); ConvT2D: two-dimensional transposed convolution.}
\label{tab:arch}
\centering
\begin{adjustbox}{max width=\textwidth}
\begin{tabular}{llll}
\toprule
 & Layer & Input & Output \\
\midrule
\multirow{5}{*}{Encoder}
 & Conv2D-1 & $1{\times}128{\times}128$ & $32{\times}64{\times}64$ \\
 & Conv2D-2 & $32{\times}64{\times}64$ & $64{\times}32{\times}32$ \\
 & Conv2D-3 & $64{\times}32{\times}32$ & $128{\times}16{\times}16$ \\
 & Conv2D-4 & $128{\times}16{\times}16$ & $256{\times}8{\times}8$ \\
 & Flatten + Linear & 16,384 & 256 \\
\midrule
\multirow{5}{*}{Decoder}
 & Linear + Reshape & 256 & $256{\times}8{\times}8$ \\
 & ConvT2D-1 & $256{\times}8{\times}8$ & $128{\times}16{\times}16$ \\
 & ConvT2D-2 & $128{\times}16{\times}16$ & $64{\times}32{\times}32$ \\
 & ConvT2D-3 & $64{\times}32{\times}32$ & $32{\times}64{\times}64$ \\
 & ConvT2D-4 & $32{\times}64{\times}64$ & $1{\times}128{\times}128$ \\
\bottomrule
\end{tabular}
\end{adjustbox}
\end{table}

\subsection{Training}
\label{sec:training}
\begin{figure}[!htbp]
\centering
\includegraphics[width=0.6\textwidth]{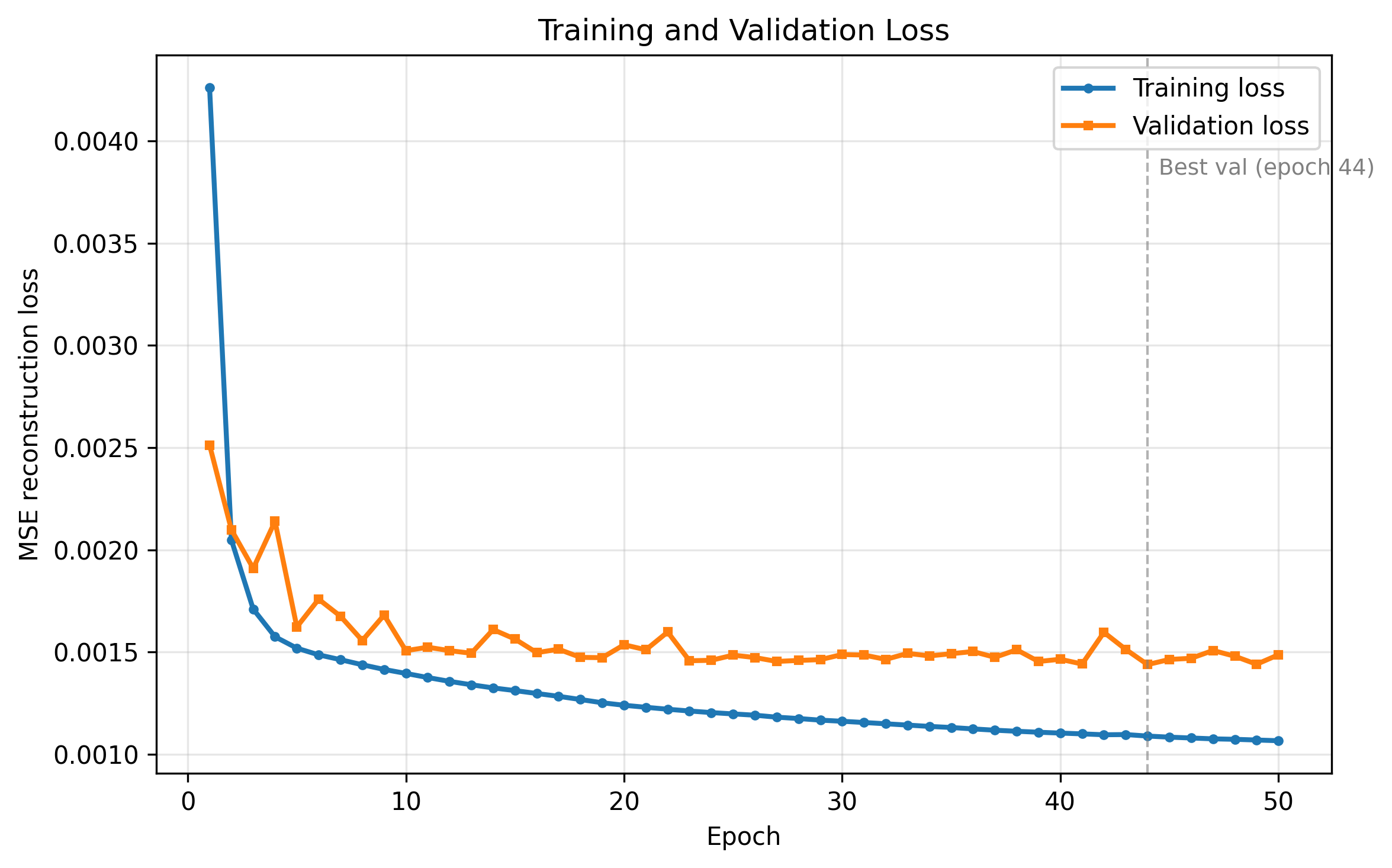}
\caption{Training and validation loss over 50 epochs. The training loss is computed on noise-corrupted inputs and the validation loss on clean ones, which contributes to the offset between them. The epoch with the lowest validation loss (44) is retained.}
\label{fig:loss}
\end{figure}

The model is trained only on normal windows. Gaussian noise ($\sigma = 0.05$) is added to each input and clipped to $[0,1]$, and the mean-squared error is computed against the clean window:
\begin{equation}
\mathcal{L} = \big\lVert x - f_\theta\!\big(\mathrm{clip}(x + \varepsilon)\big) \big\rVert_2^2, \qquad \varepsilon \sim \mathcal{N}(0, \sigma^2 I).
\end{equation}
The only augmentation is a horizontal flip with probability 0.5; a flipped B-scan corresponds to surveying the same line in the opposite direction and is therefore a physically valid sample. Each training window is presented ten times per epoch with independent flips and noise. Optimisation uses Adam \cite{kingma2015} with a learning rate of $5{\times}10^{-4}$ and batch size 64 for 50 epochs. Both losses fall rapidly in the first ten epochs and then level off without upward drift (Fig.~\ref{fig:loss}); the checkpoint with the lowest validation loss (epoch 44) is used.

\subsection{Anomaly scoring and the depth band}
\label{sec:depth}
For a window $x$ with reconstruction $\hat{x}$ the squared-error map is $E = (x - \hat{x})^2 \in \mathbb{R}^{128\times128}$. Three ways of reducing $E$ to a score are compared.

\emph{Global average.} The mean over all pixels is the usual choice, but a tunnel hyperbola occupies a small part of the window, so averaging dilutes it with well-reconstructed background.

\emph{Top-$k$ pooling.} The score averages only the largest $k\%$ of the errors,
\begin{equation}
s_{k}(x) = \frac{1}{|T_k|} \sum_{(i,j)\in T_k} E_{ij},
\end{equation}
where $T_k$ indexes the $k\%$ largest entries of $E$; $k = 5$ unless stated otherwise.

\emph{Depth band.} The vertical axis of a radargram is two-way travel time. The upper part of each window holds the direct wave and the near-surface soil returns, which change from survey to survey with moisture and surface conditions and are never reconstructed perfectly. A tunnel, however, can only appear at the depth at which it lies. The tunnel roof reflections lie at 20--33\,ns, i.e.\ rows 64--106 of a window. The depth-restricted score therefore draws $T_k$ only from rows 64--127 (20--40\,ns, about 0.85--1.7\,m):
\begin{equation}
s^{\mathrm{depth}}_{k}(x) = \frac{1}{|T_k|} \sum_{(i,j)\in T_k,\; i \geq 64} E_{ij},
\end{equation}
where $T_k$ is now the set of the $k\%$ largest entries within those rows. The band is fixed in advance and applied identically to every window; it is not learned. It removes the most variable part of the window from the score of normal ground, which tightens the threshold, and it concentrates the pooled errors of tunnel windows where the anomaly lies. Here the band was set from the depth of the tunnel roof measured on the radargrams. In operation it would follow from the depth range of concern---a tunnel aimed at a buried pipeline has to reach the depth of that pipeline---and Section~\ref{sec:robust} shows that the results change little over a wide range of band starts.

\subsection{Threshold}
\label{sec:threshold}
A window is flagged when its score exceeds
\begin{equation}
\tau = \mu_{\mathrm{normal}} + 2.5\,\sigma_{\mathrm{normal}},
\end{equation}
where $\mu$ and $\sigma$ are the mean and standard deviation of the scores of normal windows \cite{chandola2009}. The choice of normal windows used for $\mu$ and $\sigma$ is described in Section~\ref{sec:protocol}.

\subsection{Baselines}
\label{sec:baselines}
All learned baselines are fitted on the same 6,797 training windows and evaluated with and without the depth band. \emph{PCA} keeps 256 components (the size of the autoencoder code) and is scored like the autoencoder, from its reconstruction-error map. \emph{OCSVM} \cite{scholkopf2001} (radial basis function kernel, $\nu = 0.05$) and \emph{isolation forest} \cite{liu2008} (300 trees) operate on 256 PCA components of the whole window or of rows 64--127; for OCSVM the components are additionally standardised, which improved it markedly. \emph{PaDiM} \cite{defard2021} fits a Gaussian per position to 100 randomly selected ResNet-18 \cite{he2016} feature channels. \emph{PatchCore} \cite{roth2022} uses WideResNet-50 \cite{zagoruyko2016} layer-2 and layer-3 features with a 2\% greedy coreset memory bank built from every tenth training window. The anomaly maps of PaDiM and PatchCore are resized to $128{\times}128$, smoothed with a Gaussian ($\sigma = 4$ pixels at the $224{\times}224$ input), and pooled with the top-5\% rule or the maximum, over the whole map or the depth band. Every method uses the same threshold rule.

To test whether the depth band alone could serve as a detector, four scores of rows 64--127 that involve \emph{no learning} are also evaluated: the mean intensity (\emph{brightness}), the standard deviation of intensity (\emph{contrast}), the mean absolute difference between neighbouring pixels (\emph{edges}), and the mean of the largest 5\% of squared differences from the average training window (\emph{template}), the simplest possible model of normal ground. Each score was oriented in the direction that maximised its area under the receiver operating characteristic curve (AUC) on the test set, which favours these rules.

\subsection{Implementation}
All experiments were run in PyTorch on a laptop with an NVIDIA GTX 1660 Ti graphics card. The analysis and evaluation scripts were developed with the assistance of an AI coding assistant (Claude, Anthropic); all scripts are publicly available (see Data availability) and were run and checked by the authors.

\section{Evaluation protocol}
\label{sec:protocol}

Consecutive windows overlap by 94--97\%, so neighbouring windows are near-duplicates. The first two elements of the protocol address this; the other two define the operational metrics and the test on new ground.

\emph{Threshold without leakage.} The normal windows that set $\tau$ must not include the window being judged \cite{kaufman2012}. We use leave-one-line-out calibration: each survey line is classified with a threshold computed from the normal windows of the other 53 lines, and predictions are pooled. This mirrors practice, where an operator calibrates on stretches of a corridor known to be clean. Calibrating on the Site~A validation windows instead was also evaluated, as was the optimistic rule of calibrating on all test normal windows, which was used in an earlier version of this work and is reported only for reference.

\emph{Uncertainty that respects spatial dependence.} Windows within a line are not independent \cite{roberts2017}, so 95\% confidence intervals are obtained from 2,000 bootstrap resamples of whole survey lines \cite{efron1993}. Differences between methods are evaluated with the same resamples (paired bootstrap) and called significant when the interval excludes zero.

\emph{Operational metrics.} Besides the window-level AUC, average precision (AP), precision, recall and F1 score, we report the number of tunnel crossings in which at least one window was flagged, the recall at a fixed 1\% false-positive rate (FPR), and isolated false-alarm events, i.e.\ runs of consecutive flagged normal windows more than 17 windows (about one window width on lines with the 0.15-m step) from the nearest tunnel window. Tunnel windows make up about 40\% of the test set, far more than in a real survey, which favours precision and F1; recall at a fixed false-positive rate and false alarms per kilometre do not depend on this share. To test whether results depend on the gain setting chosen per line, the AUC is also computed within each line that contains both classes. Further experiments repeat training with two other random seeds, change one design choice at a time, vary the depth band and $k$, reduce the training data, and score the unseen Sites~B and C.

\emph{New ground.} To separate the effect of recalibrating the threshold from that of retraining the model, the 136 survey lines of Site~B were divided into five groups by taking every fifth line. In each of five rounds, one group served as test lines, the next as calibration lines that set the thresholds, and the remaining three were used to train a new autoencoder on Site~B's own ground (same architecture and settings, training seed 7, same number of training steps). Over the five rounds every Site~B window is a test window exactly once, and its score from the Site~A model and from the Site~B model is compared with a threshold from that round's calibration lines.

\section{Results}
\label{sec:results}

\subsection{Effect of the depth band}
Table~\ref{tab:depth} compares each method with and without the depth band. For the autoencoder, the global average reaches an AUC of 0.942 and misses 267 of 634 tunnel windows; top-5\% pooling over the whole window raises the AUC to 0.986, and restricting it to the depth band raises it to 0.994 and reduces the missed windows from 73 to 18. The depth band improves every method, often dramatically: the AUC of PatchCore rises from 0.768 to 0.926 and that of the standardised OCSVM from 0.904 to 0.993. With whole-window scores, several baselines detect almost nothing at the common threshold (F1 near or equal to zero) although their AUC is moderate, because the variable upper part of the window inflates the spread of normal scores.

\begin{table}[!htbp]
\caption{Whole window versus depth band (rows 64--127) for each method. Leave-one-line-out threshold; 634 tunnel and 966 normal windows.}
\label{tab:depth}
\centering
\small
\begin{adjustbox}{max width=\textwidth}
\begin{tabular}{llccccc}
\toprule
 & & \multicolumn{2}{c}{Whole window} & \multicolumn{3}{c}{Depth band} \\
\cmidrule(lr){3-4}\cmidrule(lr){5-7}
Method & Pooling & AUC & F1 & AUC & F1 & Missed \\
\midrule
Autoencoder (ours) & top 5\% & 0.9855 & 0.9204 & \textbf{0.9936} & \textbf{0.9716} & \textbf{18} \\
Autoencoder & global mean & 0.9417 & 0.7182 & --- & --- & --- \\
PCA & top 5\% & 0.9865 & 0.9001 & 0.9897 & 0.9303 & 73 \\
OCSVM (standardised) & --- & 0.9043 & 0.0000$^{a}$ & 0.9930 & 0.9394 & 68 \\
OCSVM & --- & 0.7396 & 0.0000$^{a}$ & 0.9614 & 0.5926 & 362 \\
Isolation forest & --- & 0.8861 & 0.4586 & 0.9904 & 0.9359 & 57 \\
PaDiM & top 5\% & 0.9347 & 0.4407 & 0.9887 & 0.9247 & 75 \\
PaDiM & maximum & 0.9057 & 0.5102 & 0.9783 & 0.8690 & 130 \\
PatchCore & top 5\% & 0.7677 & 0.0477 & 0.9259 & 0.6235 & 345 \\
PatchCore & maximum & 0.7406 & 0.0390 & 0.9063 & 0.5752 & 374 \\
\bottomrule
\multicolumn{7}{l}{\footnotesize $^{a}$No window exceeded the threshold.}
\end{tabular}
\end{adjustbox}
\end{table}

\subsection{Is a learned model of normal ground needed?}
If the depth band did all the work, a simple statistic of the band should detect the tunnel. Table~\ref{tab:nolearn} shows that it does not. The brightness of the band is close to chance (AUC 0.547) and finds one of the 29 crossings, so the tunnel does not simply make the band brighter. The strongest rule, the difference from the average training window, reaches an AUC of 0.956 but recalls only 46\% of tunnel windows at a 1\% false-positive rate and finds 20 of the 29 crossings, against 97\% and 29 for the autoencoder. The band tells the detector where to look; the learned model of normal ground is what recognises the tunnel there.

\begin{table}[!htbp]
\caption{Scores of the depth band (rows 64--127) that involve no learning, compared with the autoencoder. Leave-one-line-out threshold; each rule oriented to maximise its AUC.}
\label{tab:nolearn}
\centering
\small
\begin{adjustbox}{max width=\textwidth}
\begin{tabular}{lcccccc}
\toprule
Score & AUC & Recall at 1\% FPR & F1 & Missed & FP & Crossings \\
\midrule
Brightness (mean intensity) & 0.5466 & 0.050 & 0.025 & 626 & 6 & 1/29 \\
Edges (neighbour differences) & 0.7291 & 0.013 & 0.015 & 629 & 10 & 1/29 \\
Contrast (intensity spread) & 0.9441 & 0.416 & 0.616 & 345 & 16 & 14/29 \\
Template (difference from average window) & 0.9563 & 0.457 & 0.722 & 263 & 23 & 20/29 \\
\midrule
Autoencoder (ours) & \textbf{0.9936} & \textbf{0.968} & \textbf{0.972} & \textbf{18} & 18 & \textbf{29/29} \\
\bottomrule
\end{tabular}
\end{adjustbox}
\end{table}

\subsection{Comparison of detectors}
\begin{figure}[!htbp]
\centering
\includegraphics[width=\textwidth]{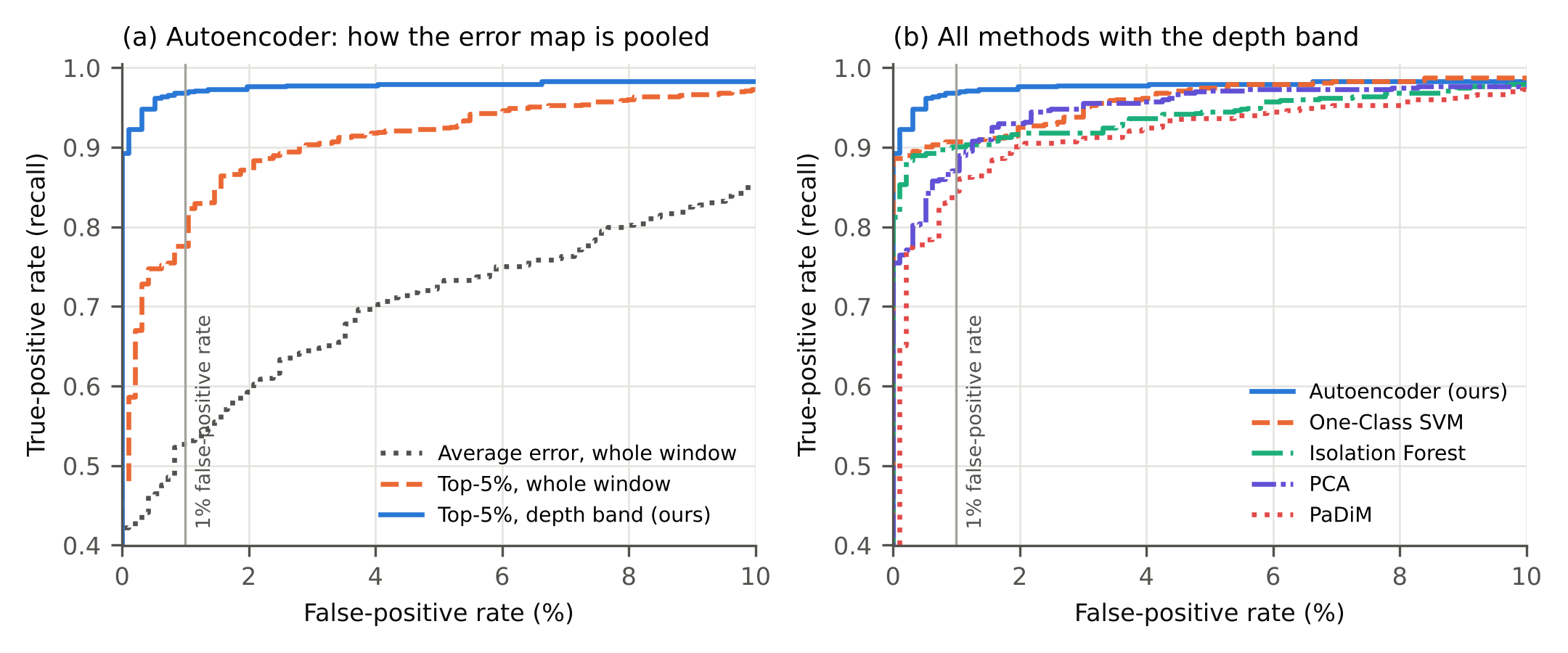}
\caption{ROC curves at low false-positive rates. (a) Autoencoder scores: global average, top-5\% over the whole window, and top-5\% within the depth band. (b) Depth-band variants of the five strongest detectors. The vertical line marks a 1\% false-positive rate.}
\label{fig:roc}
\end{figure}

Table~\ref{tab:main} compares the depth-band variants of all six detectors. The autoencoder has the highest AUC, F1 and recall, and, like the isolation forest and PCA, flags all 29 tunnel crossings. Its AUC is statistically indistinguishable from those of the standardised OCSVM ($\Delta$AUC $= +0.0006$, 95\% CI $[-0.0023, +0.0031]$), the isolation forest and PaDiM, and significantly higher than that of PCA ($+0.0039$ $[+0.0005, +0.0071]$) and PatchCore ($+0.068$ $[+0.027, +0.116]$). The differences appear at the operating points that matter in practice (Fig.~\ref{fig:roc}). At a 1\% false-positive rate the autoencoder recalls 96.8\% of tunnel windows, against 85--91\% for the OCSVM, isolation forest, PCA and PaDiM and 55\% for PatchCore. The paired 95\% intervals of these differences exclude zero for every method (e.g.\ $+6.2$ percentage points, $[+0.3, +13.5]$, against the OCSVM); after a Bonferroni correction for the five comparisons, the advantage over the OCSVM and PaDiM is no longer significant. At the common threshold the autoencoder misses 18 tunnel windows, against 57--75 for the OCSVM, isolation forest, PCA and PaDiM; its F1 advantage is significant against the isolation forest, PCA and PatchCore, but not against the OCSVM and PaDiM. The price is somewhat more false alarms than the most conservative baselines (18 windows, against 5 for the OCSVM).

\begin{table}[!htbp]
\caption{Depth-band detectors on the Site A test set. Leave-one-line-out threshold; 95\% confidence intervals from resampling whole survey lines.}
\label{tab:main}
\centering
\small
\setlength{\tabcolsep}{3.5pt}
\begin{adjustbox}{max width=\textwidth}
\begin{tabular}{lcccccccc}
\toprule
Method & AUC [95\% CI] & AP & Prec. & Recall & F1 [95\% CI] & Missed & FP & Crossings \\
\midrule
Autoencoder (ours) & \textbf{0.9936} [0.987, 0.999] & \textbf{0.9937} & 0.972 & \textbf{0.972} & \textbf{0.972} [0.933, 0.993] & \textbf{18} & 18 & \textbf{29/29} \\
OCSVM (std.) & 0.9930 [0.986, 0.998] & 0.9919 & \textbf{0.991} & 0.893 & 0.939 [0.888, 0.971] & 68 & \textbf{5} & 27/29 \\
Isolation forest & 0.9904 [0.981, 0.997] & 0.9890 & 0.963 & 0.910 & 0.936 [0.883, 0.971] & 57 & 22 & 29/29 \\
PCA & 0.9897 [0.982, 0.996] & 0.9883 & 0.981 & 0.885 & 0.930 [0.884, 0.962] & 73 & 11 & 29/29 \\
PaDiM & 0.9887 [0.981, 0.995] & 0.9845 & 0.972 & 0.882 & 0.925 [0.865, 0.963] & 75 & 16 & 27/29 \\
PatchCore & 0.9259 [0.879, 0.963] & 0.9041 & 0.986 & 0.456 & 0.624 [0.449, 0.760] & 345 & 4 & 17/29 \\
\bottomrule
\end{tabular}
\end{adjustbox}
\end{table}

\subsection{Tunnel crossings and missed windows}
\begin{figure}[!htbp]
\centering
\includegraphics[width=0.68\textwidth]{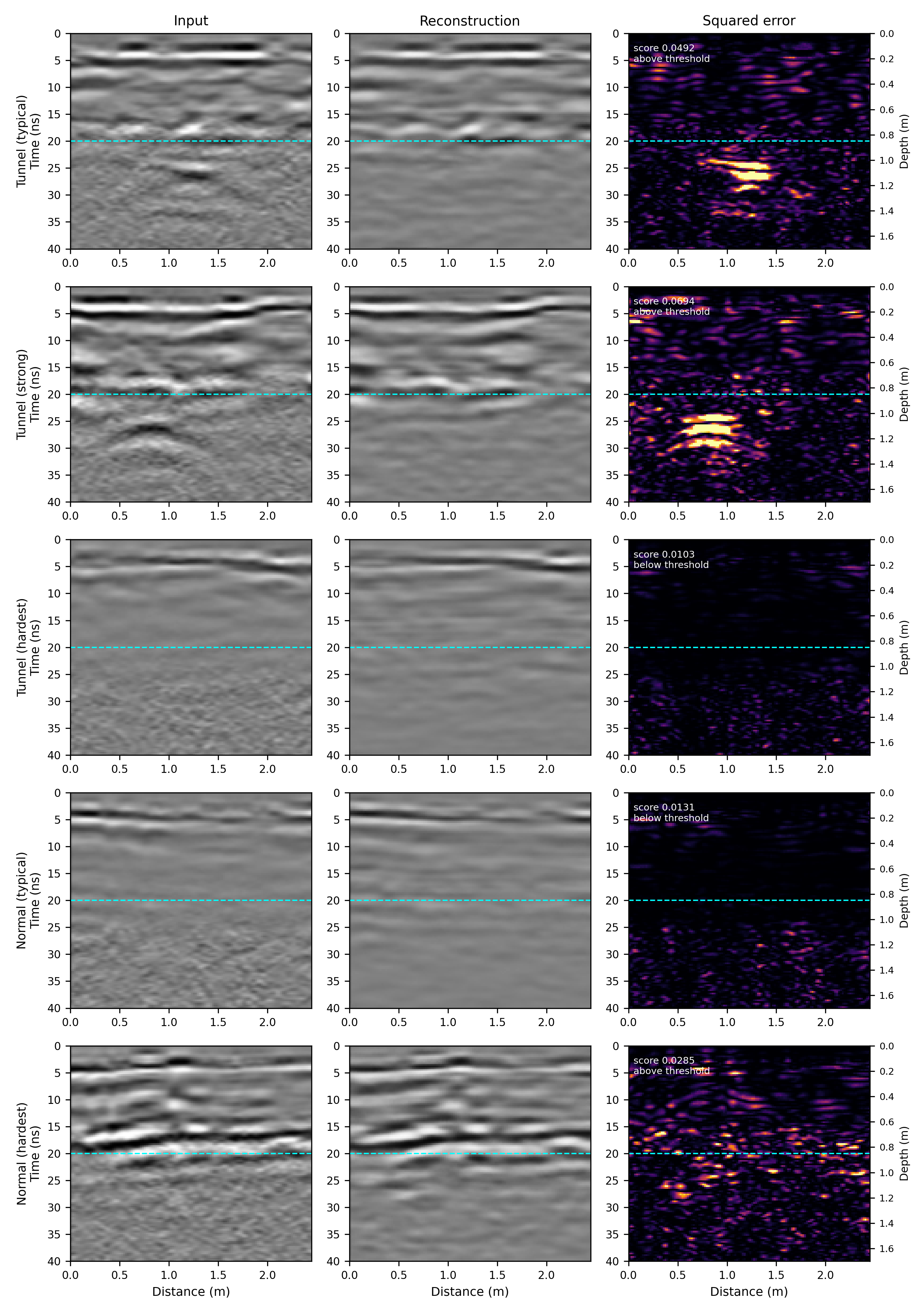}
\caption{Input, reconstruction and squared-error map for five test windows, chosen by score: the median-scoring tunnel window (typical), a tunnel window at the 90th percentile of tunnel scores (strong), the lowest-scoring tunnel window (hardest; the first tunnel window of a crossing on Line~10, where the hyperbola has barely entered the frame), the median-scoring normal window (typical), and the highest-scoring normal window (hardest; a false alarm). The dashed line marks the start of the depth band (20\,ns); the right axis gives depth at 0.085\,m/ns.}
\label{fig:errmaps}
\end{figure}

The tunnel was flagged in all 29 crossings, typically in all or nearly all of its windows. Every one of the 18 missed windows is among the first or last three windows of a crossing, where the hyperbola has only partly entered or already left the frame. One isolated window labelled \emph{tunnel} on Line~54, separated from its crossing by one window labelled normal and showing the fading hyperbola tail, was not flagged; counted as a separate run of tunnel windows it is the only one of 30 runs not detected. Fig.~\ref{fig:errmaps} shows how the error concentrates at the hyperbola apex at about 25\,ns ($\approx$1.05\,m) for tunnel windows, while normal ground reconstructs closely; the lowest-scoring tunnel window is the first window labelled \emph{tunnel} in its crossing, where the hyperbola flank has barely entered the frame; it scores the same as the normal windows before it, and the flank becomes clearly visible only in the following windows.

\begin{figure}[!htbp]
\centering
\includegraphics[width=0.75\textwidth]{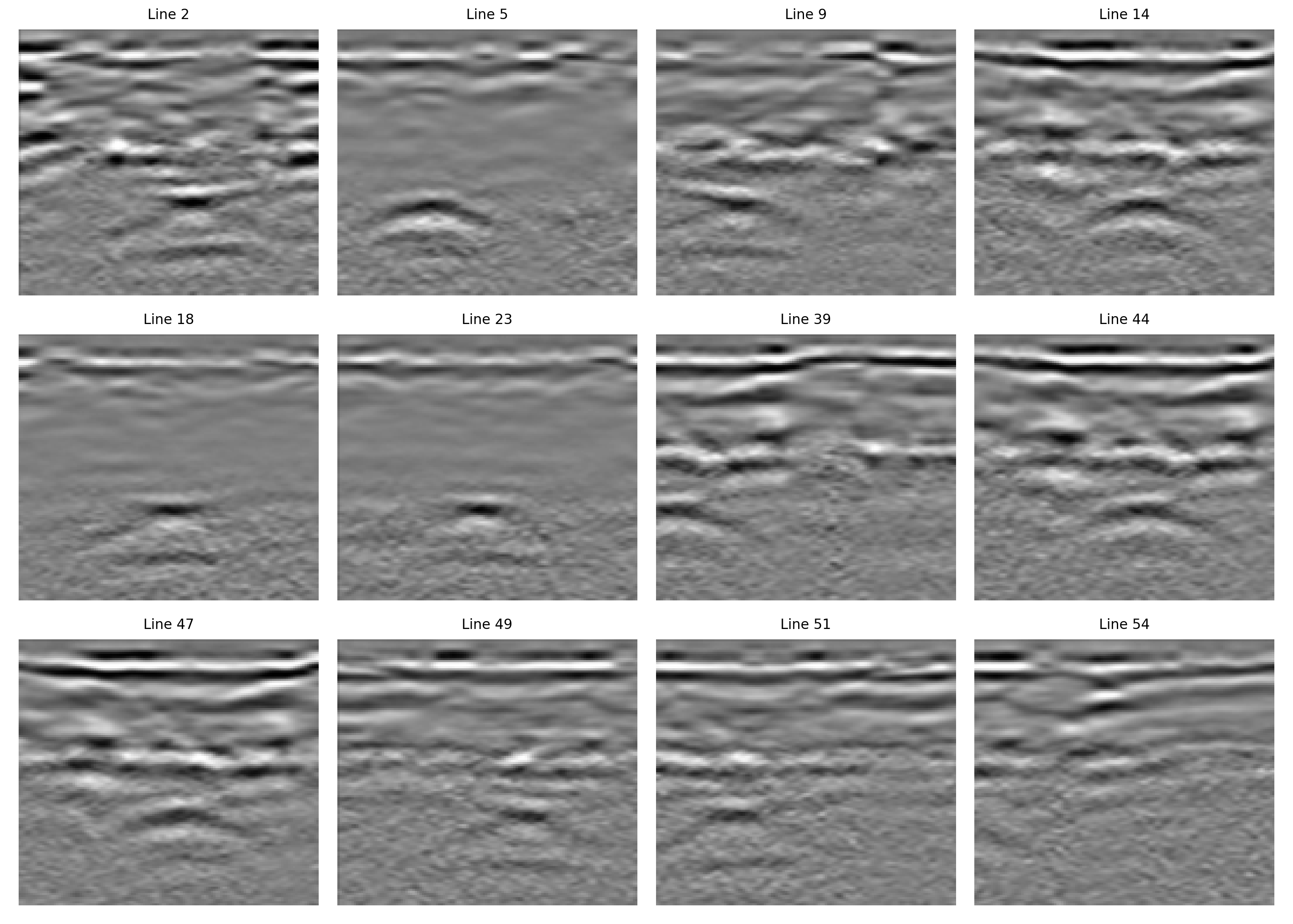}
\caption{One tunnel window from each of twelve crossings of the same tunnel, recorded in different surveys. Each panel is the middle window, in survey order, of the windows labelled \emph{tunnel} in that crossing; because a hyperbola flank often stays visible longer on one side, the apex is not always centred.}
\label{fig:same}
\end{figure}

As Fig.~\ref{fig:same} shows, the appearance of the tunnel changes considerably between surveys. In some passes it produces a sharp, compact hyperbola beneath quiet near-surface layers; in others, a faint, smeared one beneath strong layering. Its apex, however, stays at about the same travel time, which is why scoring a fixed depth band remains effective across surveys.

\subsection{False alarms}
The autoencoder flags 18 of 966 normal test windows (1.9\%). They form six events, of which one lies next to the tunnel and five are isolated. Over the roughly 115\,m of normal ground in the test set this corresponds to about 43 isolated events per kilometre; the small amount of normal ground makes this figure uncertain. The highest-scoring false alarm shows a strong, laterally continuous reflector at about 15--22\,ns (0.7--0.9\,m), just at the upper edge of the depth band, a pattern that is rare in the training surveys (Fig.~\ref{fig:soil}b) and dominates the false alarms on unseen ground (Section~\ref{sec:unseen}).

\subsection{Threshold calibration and sensitivity}
Table~\ref{tab:threshold} shows that the choice of calibration windows matters more than the rule itself. Leave-one-line-out, a 99th-percentile rule and grouped five-fold calibration all give F1 between 0.970 and 0.972. Calibrating on the pre-construction validation windows produces 231 false alarms: those windows come from the training surveys and score lower than later surveys of the same ground, so the threshold is set too low. The optimistic rule used in our earlier version (F1 0.975) is only slightly better than the line-held-out protocols. Varying $k$ in $\tau = \mu + k\sigma$ from 2.0 to 3.5 keeps F1 between 0.957 and 0.974 and all 29 crossings detected.

\begin{table}[!htbp]
\caption{Threshold calibration for the depth-band autoencoder.}
\label{tab:threshold}
\centering
\small
\begin{adjustbox}{max width=\textwidth}
\begin{tabular}{llccccc}
\toprule
Normal windows used & Rule & Prec. & Recall & F1 & Missed & FP \\
\midrule
Other survey lines (headline) & $\mu + 2.5\sigma$ & 0.972 & 0.972 & 0.972 & 18 & 18 \\
Other survey lines & 99th percentile & 0.972 & 0.968 & 0.970 & 20 & 18 \\
Other groups of lines (5-fold) & $\mu + 2.5\sigma$ & 0.972 & 0.973 & 0.972 & 17 & 18 \\
Pre-construction validation set & $\mu + 2.5\sigma$ & 0.731 & 0.991 & 0.841 & 6 & 231 \\
All test normals (reference only) & $\mu + 2.5\sigma$ & 0.976 & 0.973 & 0.975 & 17 & 15 \\
\midrule
Other survey lines & $\mu + 2.0\sigma$ & 0.941 & 0.978 & 0.959 & 14 & 39 \\
Other survey lines & $\mu + 3.0\sigma$ & 0.986 & 0.962 & 0.974 & 24 & 9 \\
Other survey lines & $\mu + 3.5\sigma$ & 0.993 & 0.923 & 0.957 & 49 & 4 \\
\bottomrule
\end{tabular}
\end{adjustbox}
\end{table}

\subsection{Robustness}
\label{sec:robust}
\emph{Processing setting.} Every survey line uses a single gain setting. Within each of the 25 lines that contain both classes, the median AUC of the autoencoder is 1.000 and 24 of the lines exceed 0.95, so the choice of setting per line cannot explain the results.

\emph{Training seed.} Retraining with two further random seeds and identical settings gives AUC $0.9937 \pm 0.0003$ and F1 $0.9717 \pm 0.0008$ (mean $\pm$ SD over three models), with all 29 crossings detected by each model.

\emph{Depth band and $k$.} Table~\ref{tab:sweep} varies the start of the band and the pooling fraction. With $k$ between 1\% and 5\%, any start between rows 32 and 80 (0.43--1.06\,m) gives an AUC of 0.991--0.995, and the choice of $k$ within that range makes little difference, so the result does not hinge on the exact band chosen. Starting at row 96 (1.28\,m), below most of the tunnel roof, lowers the AUC to about 0.956, as expected when the band no longer contains the apex; PCA shows the same drop.

\begin{table}[!htbp]
\caption{AUC of the autoencoder for different depth bands and pooling fractions.}
\label{tab:sweep}
\centering
\small
\begin{adjustbox}{max width=\textwidth}
\begin{tabular}{lcccc}
\toprule
Rows used (depth) & top 1\% & top 2\% & top 5\% & top 10\% \\
\midrule
0--127 (0--1.70\,m, whole window) & 0.9907 & 0.9902 & 0.9855 & 0.9768 \\
32--127 (0.43--1.70\,m) & 0.9953 & 0.9943 & 0.9913 & 0.9849 \\
48--127 (0.64--1.70\,m) & 0.9950 & 0.9944 & 0.9926 & 0.9891 \\
64--127 (0.85--1.70\,m, used) & 0.9949 & 0.9946 & 0.9936 & 0.9923 \\
80--127 (1.06--1.70\,m) & 0.9949 & 0.9945 & 0.9934 & 0.9924 \\
96--127 (1.28--1.70\,m) & 0.9557 & 0.9578 & 0.9561 & 0.9522 \\
\bottomrule
\end{tabular}
\end{adjustbox}
\end{table}

\subsection{Ablation}
Table~\ref{tab:ablation} changes one design choice at a time (training seed 7). Removing the input noise leaves the AUC unchanged and lowers the F1 by 0.003 (three more missed windows), so the denoising objective contributes little here. Removing the horizontal flip doubles the missed windows (16 to 32). The bottleneck size matters most: a 128-dimensional code misses 41 windows and a 512-dimensional code misses 85, consistent with a wider code also learning to reproduce the tunnel.

\begin{table}[!htbp]
\caption{Ablation of training choices (training seed 7; depth-band top-5\% score, leave-one-line-out threshold).}
\label{tab:ablation}
\centering
\small
\begin{adjustbox}{max width=\textwidth}
\begin{tabular}{lccccc}
\toprule
Model & AUC & F1 & Missed & FP & Crossings \\
\midrule
Full model & 0.9935 & 0.971 & 16 & 21 & 29/29 \\
Without input noise & 0.9937 & 0.968 & 19 & 22 & 29/29 \\
Without horizontal flip & 0.9921 & 0.959 & 32 & 19 & 29/29 \\
Bottleneck 128 instead of 256 & 0.9921 & 0.952 & 41 & 19 & 29/29 \\
Bottleneck 512 instead of 256 & 0.9857 & 0.921 & 85 & 9 & 29/29 \\
\bottomrule
\end{tabular}
\end{adjustbox}
\end{table}

\subsection{Amount of training data}
\begin{figure}[!htbp]
\centering
\includegraphics[width=0.55\textwidth]{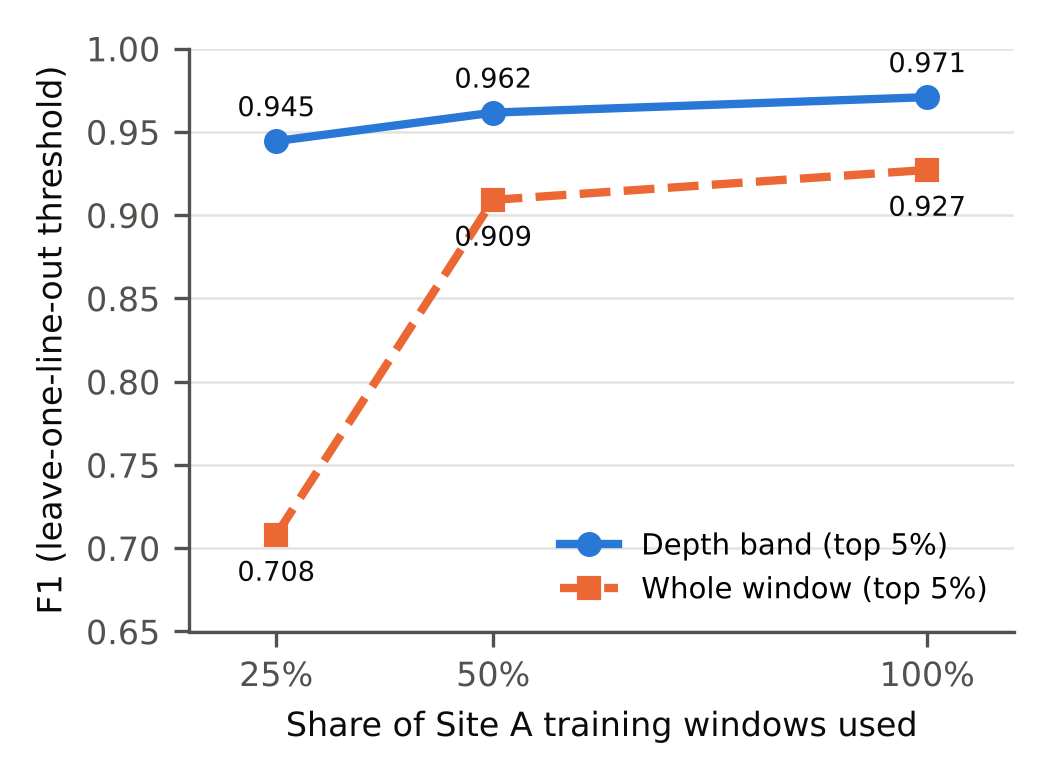}
\caption{F1 against the share of training data used, for depth-band and whole-window scoring (training seed 7).}
\label{fig:datasize}
\end{figure}

To test whether the depth band can be replaced by more data, the autoencoder was retrained on 25\% and 50\% of the Site~A training windows, selected as blocks of consecutive windows and trained for the same number of steps (Fig.~\ref{fig:datasize}). With depth-band scoring the F1 is 0.945 with 25\% of the data, 0.962 with 50\% and 0.971 with all of it. With whole-window scoring it is 0.708, 0.909 and 0.927. With a quarter of the data, the depth band outperforms whole-window scoring trained on all of it. The smaller training sets reached their lowest validation loss early (epochs 3 and 7) as the models began to memorise the repeated windows; the best-validation checkpoint was used throughout.

\subsection{Unseen ground}
\label{sec:unseen}
\begin{figure}[!htbp]
\centering
\includegraphics[width=\textwidth]{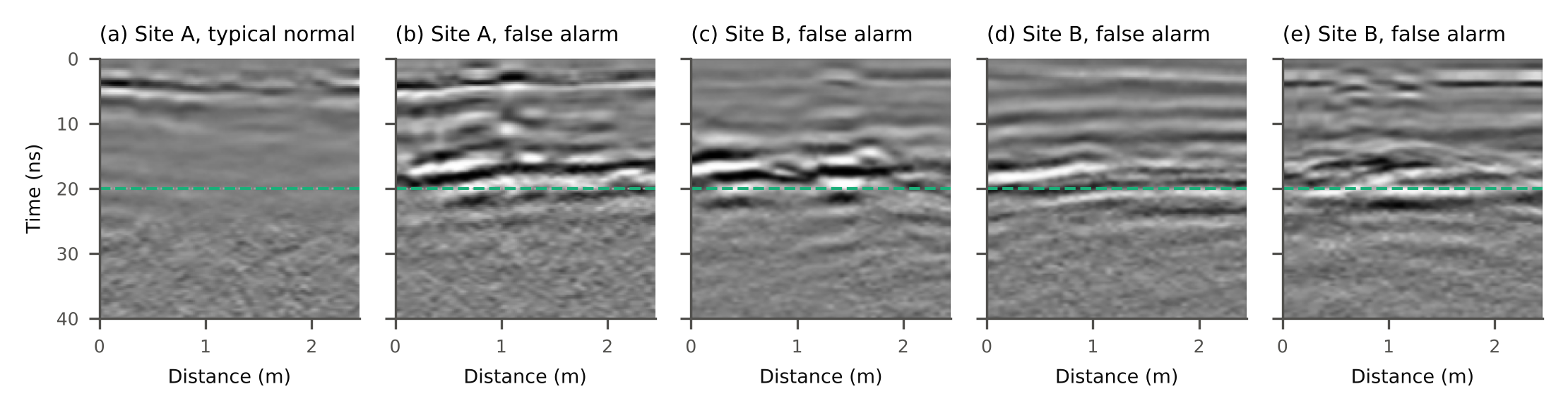}
\caption{Normal ground: (a) a typical Site~A window, (b) a Site~A false alarm, and (c--e) Site~B false alarms. The dashed line marks the start of the depth band (20\,ns). The false alarms share a strong, continuous reflector at about 15--22\,ns that is rare in the Site~A training data.}
\label{fig:soil}
\end{figure}

Table~\ref{tab:unseen} applies each detector, with the threshold from Site~A, to the 10,858 normal windows of Sites~B and C. On Site~C the autoencoder flags 1.6\% of windows, the same rate as on the normal test windows of Site~A. On the irrigated soil of Site~B the rate rises to 9.3\%. In the flagged windows we inspected, these false alarms coincide with the same strong reflector at 0.7--0.9\,m described above, which is much more common at Site~B (Fig.~\ref{fig:soil}); we attribute it to the moisture contrast between the irrigated layer and the drier soil below. When the threshold is set from Site~B's own normal lines (leave-one-line-out within the site), the rate falls to 3.5\%. Whole-window scores are less affected by this reflector (2.2\% at Site~B) but detect the Site~A tunnel far less reliably (Table~\ref{tab:depth}); the depth band trades sensitivity to deep layers on unfamiliar ground for detection on the monitored site.

\begin{table}[!htbp]
\caption{Share of normal windows flagged on the training site and on two unseen plots. Threshold from the Site~A test normal windows unless stated otherwise.}
\label{tab:unseen}
\centering
\small
\begin{adjustbox}{max width=\textwidth}
\begin{tabular}{lccccc}
\toprule
 & Site A & Site C & Site B & \multicolumn{2}{c}{Threshold from own site} \\
\cmidrule(lr){5-6}
Method (depth band unless noted) & (966) & (5,545) & (5,313) & Site C & Site B \\
\midrule
Autoencoder (ours) & 1.6\% & 1.6\% & 9.3\% & 1.7\% & 3.5\% \\
Autoencoder, whole window & 2.2\% & 0.7\% & 2.2\% & 1.9\% & 3.2\% \\
OCSVM (standardised) & 0.3\% & 0.5\% & 10.6\% & 2.3\% & 1.2\% \\
Isolation forest & 1.4\% & 0.7\% & 15.3\% & 2.7\% & 2.9\% \\
PCA & 1.0\% & 0.8\% & 6.0\% & 0.8\% & 1.4\% \\
PaDiM & 1.6\% & 3.2\% & 3.5\% & 2.9\% & 2.5\% \\
\bottomrule
\end{tabular}
\end{adjustbox}
\end{table}

\emph{Recalibrate or retrain?} Table~\ref{tab:newground} separates the two remedies using the five-round design of Section~\ref{sec:protocol}, in which every Site~B window is judged by thresholds from lines it was not part of. Recalibrating the threshold of the Site~A model on one fifth of Site~B's lines lowers the false alarms from 9.3\% to 4.2\%. Training a new autoencoder on Site~B's own ground lowers them only to 3.6\%; the difference of 0.6 percentage points is not significant (95\% CI $[-1.4, +1.8]$), and the retrained model is better on 10 lines, worse on 4 and identical on the other 122. The five Site~B models still detect the Site~A tunnel (AUC $0.988 \pm 0.003$), although less well than the Site~A model (F1 $0.940 \pm 0.007$ against 0.972), which is partly explained by their smaller training sets (2,102--3,080 windows; compare the reduced-data results above). Likewise, adding the normal windows of Site~C to the training data left detection at Site~A unchanged (F1 0.969 against 0.971, same seed) and barely changed the false-alarm rate at Site~B (8.1\% against 8.4\%). For new ground, therefore, a short baseline survey for recalibration is what matters; retraining adds little.

\begin{table}[!htbp]
\caption{Recalibrating versus retraining on unseen, irrigated ground (Site~B, all 136 survey lines, 5,313 windows; five rounds, each window tested once). 95\% confidence intervals from resampling whole survey lines.}
\label{tab:newground}
\centering
\small
\begin{adjustbox}{max width=\textwidth}
\begin{tabular}{llcc}
\toprule
Model & Threshold set from & Windows flagged & False-alarm rate [95\% CI] \\
\midrule
Site~A model & Site~A (as in Table~\ref{tab:unseen}) & 492 & 9.3\% [5.6, 12.3] \\
Site~A model & Site~B calibration lines & 224 & 4.2\% [1.6, 6.7] \\
Site~B model (retrained) & Site~B calibration lines & 193 & 3.6\% [1.2, 6.2] \\
\bottomrule
\end{tabular}
\end{adjustbox}
\end{table}

\subsection{Computational cost}
The model has 9.78 million parameters (39\,MB). On a laptop CPU (AMD Ryzen 7 4800H) a single window takes 31\,ms; on its GPU (NVIDIA GTX 1660 Ti) a batch of 64 takes 0.30\,ms per window. Even on the CPU, one window at a time, the detector processes about 17\,km of survey line per hour at the 0.15-m window step, well above the speed at which a GPR cart is pushed. Training for 50 epochs took under an hour on the same GPU when the training windows were held in memory.

\section{Discussion}
\label{sec:discussion}

\emph{What drives the false alarms.} Table~\ref{tab:variation} breaks down the scores of normal ground. On a logarithmic scale, 63--79\% of the variance of the normal Site~A test scores lies between survey lines rather than between windows of the same line: each survey is shifted as a whole, consistent with changes in moisture and surface conditions between surveys. With depth-band scoring, the pre-construction validation windows score 0.75 standard deviations below the normal test windows, which is why a threshold set from them produced 231 false alarms (Table~\ref{tab:threshold}), whereas thresholds set from other lines of the post-construction surveys did not. Unseen ground behaves in the same way. At Site~B the depth-band scores of normal windows are shifted upwards by 0.62 standard deviations while their spread changes little (a factor of 1.08); a shift of this kind is what recalibrating the threshold corrects, which is consistent with recalibration more than halving the false alarms there while retraining added little (Table~\ref{tab:newground}). At Site~C the shift is close to zero ($-0.05$), and the threshold transferred without recalibration. The depth band also separates tunnel windows more clearly: the median tunnel score lies 5.1 standard deviations above the mean normal score, against 3.0 for the whole window and 2.1 for the global average.

\begin{table}[!htbp]
\caption{Variation of normal-ground scores and separation of tunnel windows for the autoencoder. Scores on a logarithmic scale. Separation: median tunnel score minus mean normal score of the Site~A test set, in units of the standard deviation (SD) of the normal scores. Between-line share: fraction of the variance of normal test scores that lies between survey lines (Lines~1 and 2 grouped). Shifts: mean score of the validation windows or of an unseen site minus the mean normal test score, in SD; spread: ratio of their SD to that of the normal test scores.}
\label{tab:variation}
\centering
\small
\begin{adjustbox}{max width=\textwidth}
\begin{tabular}{lccccc}
\toprule
 & & & \multicolumn{3}{c}{Shift (spread) of normal scores} \\
\cmidrule(lr){4-6}
Pooling & Separation & Between-line share & Validation windows & Site B & Site C \\
\midrule
Global mean & 2.1 & 0.79 & $-0.73$ (0.42) & $+0.10$ (0.68) & $-0.16$ (0.51) \\
Top 5\%, whole window & 3.0 & 0.66 & $-0.88$ (0.50) & $+0.10$ (0.81) & $-0.28$ (0.62) \\
Top 5\%, depth band & \textbf{5.1} & 0.63 & $-0.75$ (0.70) & $+0.62$ (1.08) & $-0.05$ (0.85) \\
\bottomrule
\end{tabular}
\end{adjustbox}
\end{table}

\emph{Why the depth band works.} In reconstruction-based detection of small targets, the way the error map is pooled matters as much as the model. Near-surface returns vary with moisture and surface conditions, so their reconstruction error is large for normal ground and differs from survey to survey; any score that includes them must tolerate that variation. Restricting the score to the depth band removes this nuisance at no labelling cost. The experiments support this reading from several sides: the band helps every detector, its benefit drops sharply once the band starts below most of the tunnel roof, and it compensates for a fourfold reduction in training data that whole-window scoring cannot. The band alone, however, is not a detector: the best statistic of the band that involves no learning still misses 9 of the 29 crossings and recalls only 46\% of tunnel windows at a 1\% false-positive rate, whereas the learned model of normal ground finds every crossing and recalls 97\%.

\emph{Choice of detector.} With the depth band, several detectors reach similar AUC. The autoencoder is preferable because it misses far fewer tunnel windows at a usable false-alarm rate and is the only detector that combines the highest recall with detection of every crossing. The two ImageNet-pretrained detectors performed worst, which suggests that natural-image features do not describe radargram texture well.

\emph{Deployment.} The results suggest a simple workflow: survey a corridor while it is known to be clean, train the detector once on those surveys, set the threshold from clean stretches of each new survey or new site, and inspect flagged runs of windows. On ground similar to the training site, the threshold transfers directly: at Site~C the false-alarm rate stays at 1.6\%, the same as at Site~A, so no recalibration is needed. On markedly different ground, such as the irrigated soil of Site~B, a short baseline survey for recalibration more than halves the false alarms (from 9.3\% to 4.2\%), and retraining the model on that ground gives no significant further benefit (3.6\%). A model trained at one site can also monitor another: autoencoders trained only on the soil of Site~B, never shown a Site~A window, detected the Site~A tunnel (AUC $0.988 \pm 0.003$, F1 $0.940 \pm 0.007$) once the threshold was set from clean lines of Site~A. Because every miss occurred at the edge of a crossing, the detector loses windows rather than tunnels; an operator reviewing flagged runs would have found every crossing.

\emph{Limitations.} All tunnel windows come from a single tunnel, about 0.6\,m wide and 1\,m high with its roof 0.9--1.4\,m deep, recorded in 29 crossings over repeated surveys in dry conditions and after rain, by day and by night, in which its reflection varied considerably (Fig.~\ref{fig:same}). The tunnels dug at Sites~B and C were deeper, with roofs at about 1.5 and 2\,m, and slightly narrower; they produced no clearly interpretable reflection in the radar data, which their greater depth and smaller size, and at Site~B the wet soil, may explain. Whether a tunnel is visible at all therefore depends on its size and depth, the soil and the radar, before any detector is applied; detection under irrigated soil could not be evaluated, only false alarms, and detection of other tunnels, depths and soil types remains to be shown. The depth band was set from the tunnel's measured depth; the sweep shows that results hold over a wide range of band starts, but in a new application the band must be set from the depth range of concern. Thresholds were calibrated on windows known to be normal, which in practice requires survey stretches known to be clean. The tunnel is shallow and small, and deeper targets would require a lower antenna frequency. Window labels rest on visual identification of the hyperbola, so where exactly a crossing starts and ends is a matter of judgement. Finally, the detector classifies windows; it does not localise tunnels in three dimensions.

\section{Conclusion}
\label{sec:conclusion}
We presented an unsupervised approach to tunnel detection in GPR that learns the normal ground of a site from surveys made before a tunnel exists. Restricting the anomaly score to the depth band in which a tunnel can occur is the key element: it improves every detector tested, compensates for a fourfold reduction in training data, and with a convolutional autoencoder detects all 29 crossings of a hand-dug tunnel at an F1 score of 0.972 under a line-held-out evaluation, whereas the best score of the same band that involves no learning misses 9 of the 29 crossings. On unseen ground, false alarms remain low where the ground resembles the training site, and on very different ground they are more than halved by recalibrating the threshold from a short baseline survey, without retraining the model; a model trained only on the soil of another site also detected the tunnel. Four practical lessons follow for unsupervised detection with field sensors: score only where the target can physically appear; keep a learned model of normal ground, because simple statistics of the same region miss crossings; calibrate the threshold on data recorded under current conditions, because normal scores shift from survey to survey; and on new ground, recalibrate the threshold before considering retraining. Future work will extend the evaluation to further tunnels, depths and soils and add survey-level localisation.

\section*{CRediT authorship contribution statement}
\textbf{Muhammad Junaid:} Conceptualization, Methodology, Investigation, Data curation, Software, Formal analysis, Visualization, Writing -- original draft. \textbf{Shoab A. Khan:} Conceptualization, Supervision, Funding acquisition, Resources, Writing -- review \& editing. \textbf{Nisar Ahmed:} Validation, Writing -- review \& editing.

\section*{Declaration of competing interest}
Shoab A. Khan is the Chief Executive Officer of CARE (Centre for Advanced Research in Engineering) Pvt. Ltd., which funded this work, and the Chancellor of Sir Syed CASE Institute of Technology, where the work was carried out. The other authors declare that they have no known competing financial interests or personal relationships that could have appeared to influence the work reported in this paper.

\section*{Funding}
This work was funded by CARE (Centre for Advanced Research in Engineering) Pvt. Ltd., Islamabad, Pakistan. Its Chief Executive Officer, Shoab A. Khan, defined the research problem and supervised the work as an author; the funder had no other role in the collection, analysis and interpretation of the data, the writing of the paper, or the decision to submit it for publication.

\section*{Data availability}
The data and code supporting this study are available as follows:
\begin{itemize}
\item GPR window dataset (Kaggle):\\ \url{https://www.kaggle.com/datasets/muhammadjunaid007/gpr-normal-and-tunnel-anomaly-dataset}
\item Code, exact training and validation splits, and trained model (GitHub):\\ \url{https://github.com/Codingcahesession/gpr-tunnel-detection}
\item Raw radar files: available from the corresponding author on reasonable request.
\end{itemize}

\section*{Declaration of generative AI and AI-assisted technologies in the manuscript preparation process}
During the preparation of this work the authors used Claude (Anthropic) to help write and check the analysis and evaluation code, to summarise the experimental results, and to draft and edit the text of the manuscript. After using this tool, the authors reviewed and edited the content as needed and take full responsibility for the content of the published article.

{\raggedright

}

\end{document}